\documentclass[letterpaper]{article} % DO NOT CHANGE THIS
\usepackage{aaai25}  % DO NOT CHANGE THIS
\usepackage{times}  % DO NOT CHANGE THIS
\usepackage{helvet}  % DO NOT CHANGE THIS
\usepackage{courier}  % DO NOT CHANGE THIS
\usepackage[hyphens]{url}  % DO NOT CHANGE THIS
\usepackage{graphicx} % DO NOT CHANGE THIS
\usepackage{natbib}  % DO NOT CHANGE THIS AND DO NOT ADD ANY OPTIONS TO IT
\usepackage{caption} % DO NOT CHANGE THIS AND DO NOT ADD ANY OPTIONS TO IT
\usepackage{algorithm}
\usepackage{algorithmic}
\usepackage{booktabs} 
\usepackage{multirow}
\usepackage{xfrac}
\usepackage{amssymb}

\usepackage{newfloat}
\usepackage{listings}
\DeclareCaptionStyle{ruled}{labelfont=normalfont,labelsep=colon,strut=off} % DO NOT CHANGE THIS
\floatstyle{ruled}
\newfloat{listing}{tb}{lst}{}
\floatname{listing}{Listing}
\title{MergeUp-augmented Semi-Weakly Supervised Learning for WSI Classification}

\author{
    Mingxi Ouyang\textsuperscript{\rm 1}\thanks{These authors contributed equally to this work.}, 
    Yuqiu Fu\textsuperscript{\rm 1}\footnotemark[1], 
    Renao Yan\textsuperscript{\rm 1}\footnotemark[1]\thanks{Corresponding author.},
    ShanShan Shi\textsuperscript{\rm 1},
    Xitong Ling\textsuperscript{\rm 1},
    Lianghui Zhu\textsuperscript{\rm 1}\footnotemark[2],
    Yonghong He\textsuperscript{\rm 1}\footnotemark[2],
    Tian Guan\textsuperscript{\rm 1}\footnotemark[2]
}
\affiliations{
    \textsuperscript{\rm 1}Tsinghua Shenzhen International Graduate School, Tsinghua University, China
}

\usepackage{bibentry}
\begin{document}

\maketitle
\author {
    % Authors
    Mingxi~Ouyang\textsuperscript{\rm 1},
    Yuqiu~Fu\textsuperscript{\rm 1},
    Renao Yan\textsuperscript{\rm 1}
}
\begin{abstract}
Recent advancements in computational pathology and artificial intelligence have significantly improved whole slide image (WSI) classification. However, the gigapixel resolution of WSIs and the scarcity of manual annotations present substantial challenges. Multiple instance learning (MIL) is a promising weakly supervised learning approach for WSI classification. Recently research revealed employing pseudo bag augmentation can encourage models to learn various data, thus bolstering models' performance. While directly inheriting the parents' labels can introduce more noise by mislabeling in training. To address this issue, we translate the WSI classification task from weakly supervised learning to semi-weakly supervised learning, termed SWS-MIL, where adaptive pseudo bag augmentation (AdaPse) is employed to assign labeled and unlabeled data based on a threshold strategy. Using the "student-teacher" pattern, we introduce a feature augmentation technique, MergeUp, which merges bags with low-priority bags to enhance inter-category information, increasing training data diversity. Experimental results on the CAMELYON-16, BRACS, and TCGA-LUNG datasets demonstrate the superiority of our method over existing state-of-the-art approaches, affirming its efficacy in WSI classification.
\end{abstract}


\section{Introduction}

In recent years, the combination of digital pathology and artificial intelligence has witnessed advanced progress. Whole slide image (WSI) classification has emerged as a fundamental clinical task for diagnosing various diseases \cite{yu2023prototypical,raciti2023clinical,zhu2023accurate,zheng2024deep}. However, WSIs pose unique challenges due to their gigapixel resolution and the memory limitations of current hardware, making them difficult to process like natural images  \cite{kumar2020whole,borowsky2020digital,komura2018machine}. Additionally, the limited availability of manual annotations necessitates the use of tiled patches from WSIs, translating the classification task from a supervised learning (SL) approach to a weakly supervised learning (WSL) approach.

\begin{figure}[t]
\centering
\includegraphics[width=\linewidth]{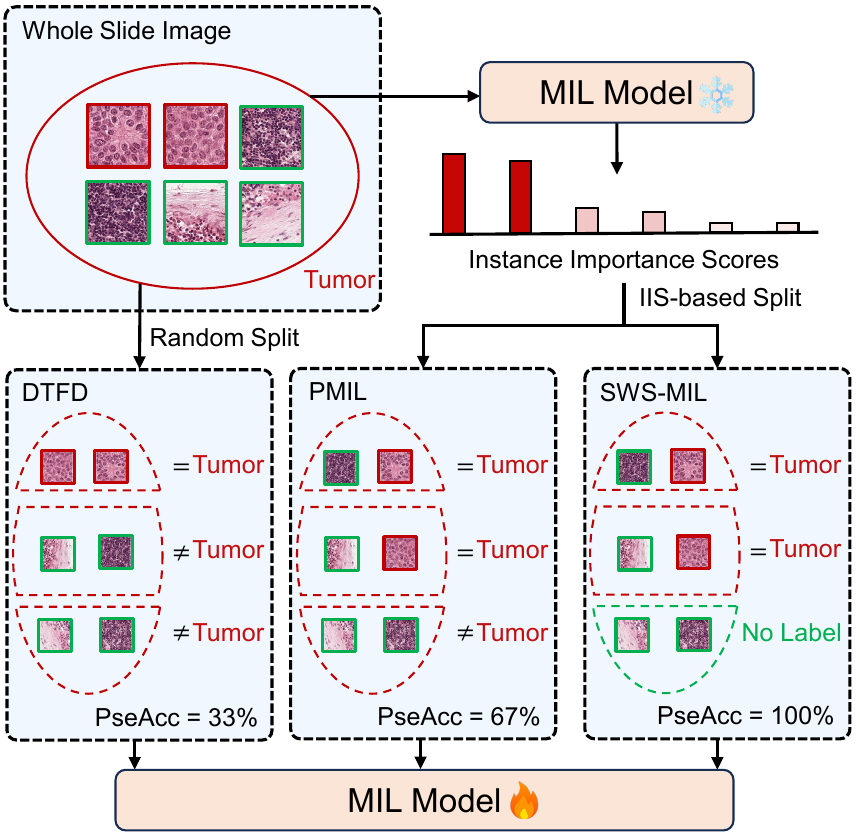}
\caption{Mislabeling issue faced by different pseudo bag augmentation strategies.}
\label{fig1}
\end{figure}

Among all the weakly supervised learning methodologies, multiple instance learning (MIL) \cite{silva2021weglenet,jia2017constrained,wang2024task} serves as a promising approach for WSI classification, where at least a positive instance in one bag marks the bag positive, otherwise negative. Current research in MIL focuses on extracting instance-level information from slide-level labels to enhance MIL training and fine-tune feature encoders. In attention-based MIL models, attention scores generated from attention pooling are used to estimate instance importance scores (IIS) \cite{yan2023shapley}. However, this approach is observed experimentally to overfitting due to the extreme distribution of attention scores.

To address this challenge, a notable strategy involves dividing regular bags into several pseudo bags, encouraging MIL models to learn from a greater variety of bags. As depicted in Figure~\ref{fig1}, in existing pseudo bag-based approaches, the random split is commonly adopted, where each pseudo bag inherits its parent label. However, as the number of pseudo bags increases, more noise is introduced as an increasing proportion of pseudo bags possess true labels that are inconsistent with their inherited labels. Note that the true labels of pseudo bags are unavailable, a common practice is to adjust the number of pseudo bags to find an optimal trade-off, aiming to maximize valid information from accurate pseudo labels while minimizing mislabeling noise \cite{yan2023shapley}. This rough strategy ties the NUMBER and LABEL assignment of pseudo bags, which is suboptimal since different WSIs have varying quantities of positive instances.

\begin{figure*}[ht]
\centering
\includegraphics[width=0.89\textwidth]{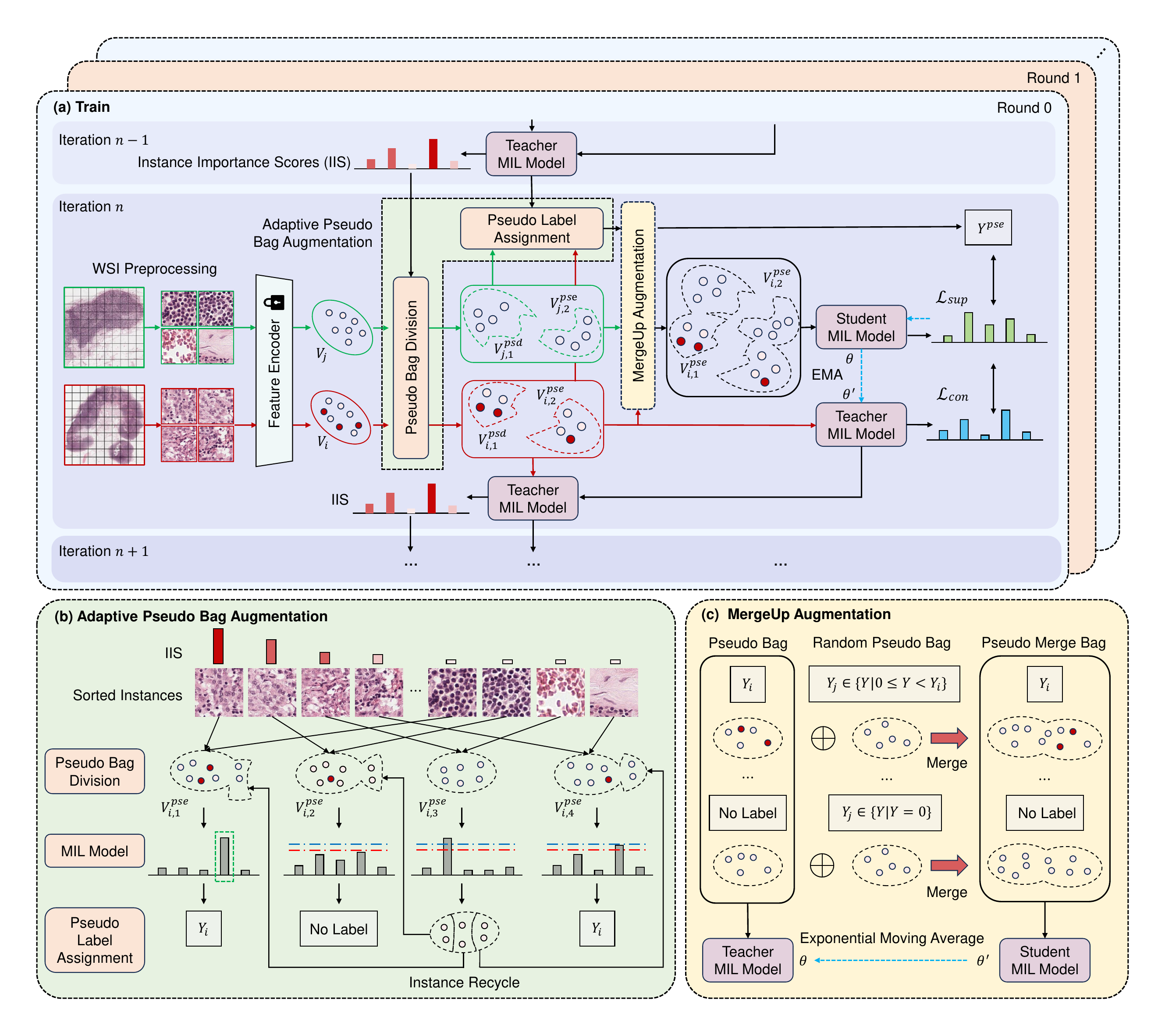}
\caption{Overview of proposed semi-weakly supervised learning framework SWS-MIL. (a) All instances in WSIs are augmented by adaptive pseudo bags for the teacher model, and MergeUp augmentation is then applied for the student model. (b) Adaptive pseudo bag augmentation determines whether the pseudo bag label is correct and recycles the incorrectly labeled instances. (c) MergeUp augmentation blends instances from low-priority pseudo bags into high-priority pseudo bags.}
\label{overflow}
\end{figure*}

To overcome this limitation, we propose an adaptive pseudo bag assignment (AdaPse) method, which separates the number assignment and label assignment of pseudo bags, achieving a better trade-off point. In this approach, a threshold-based strategy is employed to filter out pseudo bags with high confidence. Pseudo bags whose predictions are inconsistent with their parent labels are discarded and recycled, while those with predictions consistent with their parent labels inherit their parent labels. The remaining pseudo bags with low confidence are left unlabeled. This method transforms the WSI classification task from a weakly supervised learning issue to a \textbf{semi-weakly supervised learning (SWSL)} issue.

Referring to semi-supervised learning (SSL), we adopt the “student-teacher” pattern \cite{tarvainen2017mean} as our main framework, applying the “weak and strong augmentation” strategy \cite{xie2020unsupervised} for consistency regularization \cite{laine2016temporal}. In computational pathology, feature augmentation is rarely studied, as most data augmentation techniques are unusable. Since the WSI classification task is based on one-hot labels, existing feature augmentation methods fail to capture the nuanced inclusion and exclusion relationships between categories, which is crucial when categories are not mutually exclusive, such as normal versus tumor, or cancer staging.

To address this issue, we propose a novel feature augmentation method, "MergeUp", which merges different category bags while retaining the high-priority label, allowing models to learn the prioritization between different categories. In summary, our main contributions are as follows:

1. We propose a novel multiple instance learning framework SWS-MIL, where adaptive pseudo bag assignment, namely AdaPse is introduced for more reasonable pseudo augmentation, translating the weakly supervised learning task into a semi-weakly supervised learning task.

2. We introduce a feature augmentation method, MergeUp, to incorporate inter-category information into MIL training, which is suitable for non-mutually exclusive tasks and significantly increases the data diversity.

3. Extensive experiments on the CAMELYON-16, BRACS, and TCGA-LUNG datasets demonstrate that our method outperforms other state-of-the-art methods.

\section{Related Works}

\subsection{Multiple Instance Learning}
Current MIL models can be categorized into two types based on whether the final bag predictions are derived from direct instances \cite{avolio2020semiproximal,gang2013medical,bernardini2020early,yang2023multiple,carbonneau2016witness} or the aggregated features of instances \cite{chen2006miles,wang2018revisiting,chen2006miles}. ABMIL \cite{ilse2018attention} employed an attention mechanism to evaluate the contribution of each instance to the bag label. CLAM \cite{lu2021data} utilized instance-level clustering to classify slides with slide-level labels. DTFD \cite{zhang2022dtfd} addressed the challenge of small sample cohorts in WSI classification by proposing pseudo bags to expand the number of bags and utilize intrinsic features. TransMIL \cite{shao2021transmil} developed transformer-based attention to leveraging morphological and spatial information, thereby improving accuracy and convergence in WSI pathology diagnosis. PMIL \cite{yan2023shapley} introduced Shapley values rather than attention scores to estimate the importance of each instance, achieving class-wise interpretation.

\subsection{Feature Augmentation} 
In deep learning, data augmentation is commonly used to improve generalization. However, most of these methods are not suitable for features under the MIL paradigm. Recently, MixUp \cite{zhang2017mixup} has been introduced to MIL due to its strong adaptability. ReMix \cite{yang2022remix} proposed a “Mix-the-Bag” augmentation, which includes reducing and mixing. RankMix \cite{chen2023rankmix} mixed ranked features from pairs of WSIs. PseMix \cite{liu2024pseudo} combined pseudo bag augmentation with the MixUp strategy. Although these methods attempted to introduce augmentation techniques from other fields into MIL, they overlooked the issue of non-mutual exclusivity between categories originating from clinical practice, resulting in soft labels from MixUp being illogical.

\subsection{Semi-supervised Learning} 
Semi-supervised learning aims to enhance models' performance by leveraging large volumes of unlabeled data, commonly employing the "student-teacher" pattern and consistency regularization. Mean Teacher \cite{tarvainen2017mean} improved performance by averaging model weights instead of label predictions. UDA \cite{xie2020unsupervised} introduced a "weak and strong augmentation" strategy for consistency regularization. FixMatch \cite{sohn2020fixmatch} used a fixed high threshold (0.95) to ensure high-quality pseudo labels. FreeMatch \cite{wang2022freematch} dynamically adjusted the confidence threshold and employed class fairness regularization to improve performance.

\section{Methods} \label{Methods}
In this section, we first review the MIL paradigm, then introduce various feature augmentation techniques, and finally propose our framework, as shown in Figure \ref{overflow}.

\subsection{MIL for WSI Classification}
In this task, the input data, namely a set of labeled WSIs $\mathcal{D}=\left\{X_i,Y_i\right\}_{i=1}^{\left|\mathcal{D}\right|}$, are always tiled into $N_i$ patches $X_i=\left\{x_{i,j}\right\}_{j=1}^{N_i}$ for feature extraction. Conventionally, each instance $x_{i,j}$ is extracted by a convolutional neural network $h\left(\cdot\right)$ to obtain the instance-level feature $v_{i,j}$:
\begin{equation}
v_{i,j} = h\left(x_{i,j}\right)
\end{equation}
where the feature encoder $h\left(\cdot\right)$ is usually frozen during training due to memory bottlenecks. All instance-level features in one bag will be aggregated for slide-level prediction $\widehat{Y_i}$ with confidence $\gamma_i$:
\begin{equation}
\widehat{Y_i}=\max \gamma_i=\max\left\{\left(f\circ g\right)\left(\left\{v_{i,j}\right\}_{j=1}^{N_i}\right)\right\},
\end{equation}
where $g\left(\cdot\right)$ and $f\left(\cdot\right)$ represent the aggregator and classifier.

\subsection{Feature Augmentation} 
During MIL training, the fixed nature of the feature encoder often leads to overfitting. To mitigate this issue, we introduce feature augmentation techniques.

\subsubsection{Adaptive Pseudo Bag.} 
One strategy to prevent the model from overfocusing on limited instances is to randomly split the bag $V_i$ into $M$ pseudo bags for training, denoted as:
\begin{equation}
V_i=\left\{v_{i,j}\right\}_{j=1}^{N_i}=V_{i,1}^{pse}\cup V_{i,2}^{pse}\cup\cdots\cup V_{i,M}^{pse}.
\end{equation}
Since the true pseudo bag label $Y_{i,a}^{pse}$ is unavailable, each pseudo bag $V_{i,a}^{pse}$ inherits the label $Y_i$ from the parent bag, assigning $\bar{Y}_{i,a}^{pse} = Y_i$. This forms a pseudo bag dataset $\mathcal{D}^{pse} = \left\{V_{i,a}^{pse}, \bar{Y}_{i,a}^{pse} = Y_i\right\}_{i=1}^{\left|\mathcal{D}\right|}, a \in \left\{1,2,\cdots,M\right\}$.

Random splitting can introduce significant label noise when $\bar{Y}_{i,a}^{pse} \neq Y_i$, which destabilizes training. To address this issue, progressively assigning pseudo bags based on IIS has been experimentally shown to be effective \cite{yan2023shapley}, where all instances in one bag are sorted in descending order of estimated IIS:
\begin{equation}
V_i^\prime=\left\{\left.v_{i,j}^\prime\right|\ {\rm IIS}\left(v_{i,1}^\prime\right)\geq {\rm IIS}\left(v_{i,2}^\prime\right)\geq\cdots\geq {\rm IIS}\left(v_{i,N_i}^\prime\right)\right\}
\end{equation}

Then these ordered instances are evenly interleaved into $M$ pseudo bags:
\begin{equation}
V_{i,a}^{pse}=\left\{v_{i,\left(k-1\right)M+a}^\prime\right\}_{k=1}^{\sfrac{N_i}{M}},a\in\left\{1,2,\cdots,M\right\}.
\end{equation}

This approach can reduce noise generated by pseudo bag mislabeling to some extent. To take full advantage of pseudo bag augmentation, the number of pseudo bags $M$ is usually set to a constant, which can be relatively large to extract more information from bags with many positive instances. However, in bags with only a few positive instances, the number of positive instances can be much lower than the given constant, where this framework cannot help.

To identify mislabeled pseudo bags, we freeze the training model's parameters to provide predictions with confidence for each pseudo bag. Among all pseudo bags whose predictions are inconsistent with their parent labels, a fixed confidence threshold $\gamma_{fix}$ is applied to filter out and discard those mislabeled pseudo bags, forming a discarded set $\mathcal{D}_{dc}^{pse}=\left\{\left. V_{i, a}^{pse},Y_i \right| \widehat{Y}_{i, a}^{pse} \neq Y_i, \gamma_{i, a}^{pse} \geq \gamma_{fix}\right\}_{i=1}^{\left|\mathcal{D}\right|},a\in\left\{1,2,\cdots\right\}$ and a remaining set $\mathcal{D}_{rm}^{pse} = \mathcal{D}^{pse} - \mathcal{D}_{dc}^{pse}$.

To reuse these discarded mislabeled pseudo bags $\left\{V_{i, b}^{pse}, V_{i, c}^{pse}, \cdots\right\} \in \mathcal{D}_{dc}^{pse}$, we typically recycle these instances to the remaining pseudo bags $V_{i, a}^{pse}\in\mathcal{D}_{rm}^{pse}$ evenly:
\begin{equation}
V_{i, b, a}^{pse}=\left\{v_{i, b, \left(k-1\right)M+a}^{pse}\right\}_{k=1}^{\sfrac{\left|\mathcal{D}_{dc}^{pse}\right|}{\left|\mathcal{D}_{rm}^{pse}\right|}},
\end{equation}
\begin{equation}
V_{i, a}^{pse\ast}=V_{i, a}^{pse}\cup \left(V_{i, b, a}^{pse}\cup V_{i, c, a}^{pse}\cup\cdots\right).
\end{equation}

Then a new pseudo bag set $\mathcal{D}^{pse\ast}=\left\{V_{i, a}^{pse\ast},Y_i \right\}_{i=1}^{\left|\mathcal{D}\right|},a\in\left\{1,2,\cdots\right\}$ is formed. For pseudo bags $V_{i, a}^{pse}\in\mathcal{D}^{pse\ast}$ whose predictions match their parent labels, an adaptive confidence threshold $\gamma_{ada}$ is applied to filter out and assign those highly confident pseudo bags their parent labels, denoted by:
\begin{equation}
\mathcal{D}_{lb}^{pse}=\left\{\left. V_{i, a}^{pse},Y_i \right| \widehat{Y}_{i, a}^{pse} = Y_i, \gamma_{i, a}^{pse} \geq \gamma_{ada}\right\}.
\end{equation}

As the training model is initially cautious in prediction, we set the threshold $\gamma_{ada}$ to a small constant and progressively increase it. The remaining pseudo bags are likely to contaminate the dataset, so they are treated unlabeled:
\begin{equation}
\mathcal{D}_{ulb}^{pse}=\left\{\left. V_{i, a}^{pse}\right| \widehat{Y}_{i, a}^{pse} = Y_i, \gamma_{i, a}^{pse} < \gamma_{ada}\right\}.
\end{equation}

Thus, the training set can be divided into a labeled pseudo bag set $\mathcal{D}_{lb}^{pse}$ and an unlabeled one $\mathcal{D}_{ulb}^{pse}$, transforming weakly supervised learning into semi-weakly supervised learning and enhancing the diversity of training data.

\subsubsection{MergeUp.} 
In computational pathology, patches, and slides are prioritized in classification. For instance, in cancer subtyping, high-graded subtypes are typically more dominating than low-graded ones. It suggests that merging a lower-priority bag retains the same slide-level label. In tumor detection, this can be interpreted as tumor bags remaining tumor regardless of how many normal bags are merged. Thus, the MergeUp augmentation can be represented as:
\begin{equation}
V_{i;j}^{mrg}=V_i\cup V_j,
\end{equation}
\begin{equation}
Y_{i;j}^{mrg}=max\left\{Y_i,Y_j\right\}.
\end{equation}
\subsection{Semi-weakly Supervised Learning}
We adopt the “student-teacher” pattern and “weak and strong augmentation” strategy in our framework. Formally, we regard identity as a “weak” augmentation for the teacher model and MergeUp as a “strong” augmentation for the student model, denoted by:
\begin{equation}
{\hat{Y}}_{i,a}^{pse,t}=\left(f^t\circ g^t\right)\left(V_{i,a}^{pse}\right),
\end{equation}
\begin{equation}
{\hat{Y}}_{i,a;j,b}^{pse,s}=\left(f^s\circ g^s\right)\left(V_{i,a}^{pse}\cup V_{j,b}^{pse}\right),
\end{equation}
where the superscripts $t$ and $s$ refer to teacher and student models. For unlabeled pseudo bags, they are trained by consistency loss between student and teacher models:
\begin{equation}
\mathcal{L}_{con}=\mathbb{E}_{f^t,g^t;f^s,g^s}\left [ \left \| {\hat{Y}}_{i,a}^{pse,t}-{\hat{Y}}_{i,a;j,b}^{pse,s}\right \|^2\right ].
\end{equation}

For labeled pseudo bags, they are trained by:
\begin{equation}
\mathcal{L}_{sup}=\mathcal{H}\left({\hat{Y}}_{i,a}^{pse,s},{\bar{Y}}_{i,a}^{pse}\right),
\end{equation}
where $\mathcal{H}\left(\cdot,\cdot\right)$ refers to the cross-entropy loss. The overall loss is then set to:
\begin{equation}
\mathcal{L}=\frac{1}{2}\mathcal{L}_{con}+\frac{1}{2}\mathcal{L}_{sup}.
\end{equation}

\section{Experiences}
\subsection{Datasets and Evaluation Metrics}
\textbf{CAMELYON-16} is designed for detecting lymph node metastases in early-stage breast cancer. It consists of 399 WSIs, partitioned into 270 images for training and 129 for testing. We follow a 3-fold cross-validation protocol within the official training set, randomly selecting 20\% as the validation set and using the remaining 80\% for training.
\begin{table*}[ht]
    \caption{Slide-level performance results on CAMELYON-16, BRACS and TCGA-LUNG. The subscripts are the standard deviation of each metric. The best evaluation results are in bold.} 
    \label{compare_with_other_MIL}
    \centering
    \resizebox{\textwidth}{!}{
    \begin{tabular}{ccccccccccc}
    \toprule
    \multirow{2}{*}{\textbf{Method}} & \multicolumn{3}{c}{\textbf{CAMELYON-16}} & \multicolumn{3}{c}{\textbf{BRACS}} & \multicolumn{3}{c}{\textbf{TCGA-LUNG}}  \\
    \cmidrule(r){2-4} \cmidrule(lr){5-7} \cmidrule(lr){8-10}
     & ACC(\%) & AUC(\%) & F1(\%) & ACC(\%) & AUC(\%) & F1(\%) & ACC(\%) & AUC(\%) & F1(\%) \\
    \midrule
    MeanMIL 
    & 71.06 \textsubscript{1.93} & 58.40 \textsubscript{1.30} & 62.52 \textsubscript{3.91} 
    & 52.41\textsubscript{2.58} & 69.17\textsubscript{1.64} & 40.64\textsubscript{2.46}
    & 82.01 \textsubscript{0.90} & 88.90 \textsubscript{2.00} & 82.00 \textsubscript{1.00} \\
    MaxMIL 
    & 83.98 \textsubscript{1.59} & 87.98 \textsubscript{0.99} & 81.75 \textsubscript{1.86} 
    & 55.86\textsubscript{2.78} & 75.88\textsubscript{1.61} & 50.29\textsubscript{4.02}
    & 88.70 \textsubscript{1.00} & 94.40 \textsubscript{1.20} & 88.70 \textsubscript{1.00} \\
    DSMIL~\cite{li2021dual} 
    & 77.52 \textsubscript{1.68} & 76.76 \textsubscript{1.27} & 74.43 \textsubscript{2.72} 
    & 53.10\textsubscript{2.20} & 70.82\textsubscript{3.30} & 46.10\textsubscript{3.71}
    & 86.20 \textsubscript{1.40} & 93.60 \textsubscript{1.00} & 86.20 \textsubscript{1.40} \\
    ABMIL~\cite{ilse2018attention} 
    & 83.21 \textsubscript{0.97} & 84.60 \textsubscript{0.55} & 81.33 \textsubscript{0.88} 
    & 58.39\textsubscript{0.86} & 76.14\textsubscript{0.64} & 54.74\textsubscript{2.29}
    & 87.60 \textsubscript{0.70} & 93.10 \textsubscript{1.80} & 87.60 \textsubscript{0.70} \\
    CLAM~\cite{lu2021data} 
    & 84.50 \textsubscript{2.19} & 83.37 \textsubscript{0.90} & 82.50 \textsubscript{2.27} 
    & 53.79\textsubscript{3.52} & 73.25\textsubscript{1.65} & 51.50\textsubscript{3.29}
    & 88.20 \textsubscript{1.40} & 94.20 \textsubscript{1.20} & 88.20 \textsubscript{1.40} \\
    TransMIL~\cite{shao2021transmil} 
    & 85.27 \textsubscript{1.10} & 88.75 \textsubscript{0.60} & 83.81 \textsubscript{0.96} 
    & 57.01\textsubscript{2.37} & 75.46\textsubscript{1.00} & 49.22\textsubscript{5.15}
    & 87.90 \textsubscript{0.80} & 94.80 \textsubscript{0.80} & 87.90 \textsubscript{0.90} \\
    DTFD ~\cite{zhang2022dtfd} 
    & 84.76 \textsubscript{1.83} & 84.82 \textsubscript{3.94} & 84.45 \textsubscript{1.93} 
    & 57.24\textsubscript{2.66} & 76.55\textsubscript{1.99} & 56.20\textsubscript{3.76}
    & 88.80 \textsubscript{0.60} & 94.60 \textsubscript{0.80} & 88.80 \textsubscript{0.60} \\
    R$^2$\text{T}~\cite{tang2024feature} 
    & 74.69 \textsubscript{2.14} & 73.53 \textsubscript{5.50} & 72.01 \textsubscript{3.26} 
    & 66.00 \textsubscript{4.78} & 69.93 \textsubscript{3.52} & 53.30 \textsubscript{3.07}
    & 88.29 \textsubscript{0.61} & 79.06 \textsubscript{0.74} & 78.99 \textsubscript{0.90} \\
    PMIL~\cite{yan2023shapley} 
    & 88.11 \textsubscript{0.97} & 89.34 \textsubscript{1.60} & 86.99 \textsubscript{0.82}
    & 68.95 \textsubscript{0.16} & 84.65 \textsubscript{1.62} & 54.32 \textsubscript{5.76}
    & 91.30 \textsubscript{1.40} & 96.50 \textsubscript{0.90} & 91.30 \textsubscript{1.40} \\
    \midrule
    MixUp~\cite{zhang2017mixup} 
    & 66.93 \textsubscript{2.72} & 75.29 \textsubscript{1.72} & 66.23 \textsubscript{2.40}
    & 30.38\textsubscript{10.46} & 55.77 \textsubscript{1.90} & 53.20 \textsubscript{5.60}
    & 66.83 \textsubscript{2.66} & 71.32 \textsubscript{2.83} & 64.96 \textsubscript{4.08} \\
     AddictiveMIL~\cite{javed2022additive} 
    & 67.27 \textsubscript{6.10} & 56.40 \textsubscript{3.40} & 59.58 \textsubscript{4.80}
    & 54.25 \textsubscript{2.60} & 74.18 \textsubscript{0.50} & 47.37 \textsubscript{1.90}
    & 80.95 \textsubscript{3.60} & 88.25 \textsubscript{1.40} & 80.83 \textsubscript{3.80} \\
    RankMix~\cite{chen2023rankmix} 
    & 66.20 \textsubscript{4.26} & 73.00 \textsubscript{4.14} & 65.58 \textsubscript{4.20}
    & 40.08 \textsubscript{4.61} & 60.09 \textsubscript{8.33} & 36.32\textsubscript{10.25}
    & 71.45 \textsubscript{1.45} & 75.93 \textsubscript{1.82} & 71.23 \textsubscript{0.90} \\
    PseMix~\cite{liu2024pseudo} 
    & 85.60 \textsubscript{3.30} & 86.10 \textsubscript{4.40} & 78.30 \textsubscript{5.50}
    & 57.90 \textsubscript{3.50} & 77.30 \textsubscript{1.90} & 53.20 \textsubscript{5.60}
    & 89.80 \textsubscript{0.60} & 95.20 \textsubscript{1.30} & 89.50 \textsubscript{0.90} \\
    \midrule
    %SWS-MIL-Random(OURS)     & 89.66 \textsubscript{1.30} & \ \textbf{94.00 \textsubscript{0.90}} & 88.43 \textsubscript{1.60} &\textbf{70.60 \textsubscript{2.10}} & \textbf{85.00 \textsubscript{1.00}} & 58.70 \textsubscript{3.00} &\textbf{92.30 \textsubscript{0.70}} & \textbf{97.20 \textsubscript{0.50}} & 92.20 \textsubscript{0.70}\\
   SWS-MIL (Shapley) 
   & 89.92 \textsubscript{1.70} & 93.04 \textsubscript{1.00} & \ 88.82 \textsubscript{2.00}
   & \textbf{69.80 \textsubscript{1.50}} & \ \textbf{84.90 \textsubscript{1.20}} & \ 58.20 \textsubscript{2.90}
   & 91.50 \textsubscript{0.60} & \textbf{97.00 \textsubscript{0.60}} & 91.40 \textsubscript{0.60}\\
   SWS-MIL (Attention) 
   & \textbf{90.18 \textsubscript{1.60}} & \textbf{93.25 \textsubscript{0.70}} & \textbf{89.27 \textsubscript{1.70}}
   & 69.40 \textsubscript{4.00} & \ 84.40 \textsubscript{1.50} & \textbf{59.70 \textsubscript{7.70}}
   & \textbf{91.60 \textsubscript{0.50}} & 96.80 \textsubscript{0.70} & \textbf{92.30 \textsubscript{0.50}}\\
    % SWS-MIL(OURS) & 88.89\textsubscript{0.45} & 92.47\textsubscript{1.07} & 87.37\textsubscript{0.60} & 75.09\textsubscript{4.04} & 87.79\textsubscript{3.69} & 68.08\textsubscript{4.84}\\
    \bottomrule
    \end{tabular}
    }
\end{table*}

\textbf{BRACS}~\cite{brancati2022bracs} is specifically curated for the nuanced task of breast cancer subtyping, serving as a critical resource for developing and evaluating automated classification systems. It comprises 547 WSIs, meticulously annotated for classification into three categories: benign tumors, atypical tumors (AT), and malignant tumors (MT). We follow a 3-fold cross-validation protocol, with the training/validation/test datasets set at a ratio of 7:1:2.

\begin{figure}[ht]
\centering
\includegraphics[width=0.95\linewidth]{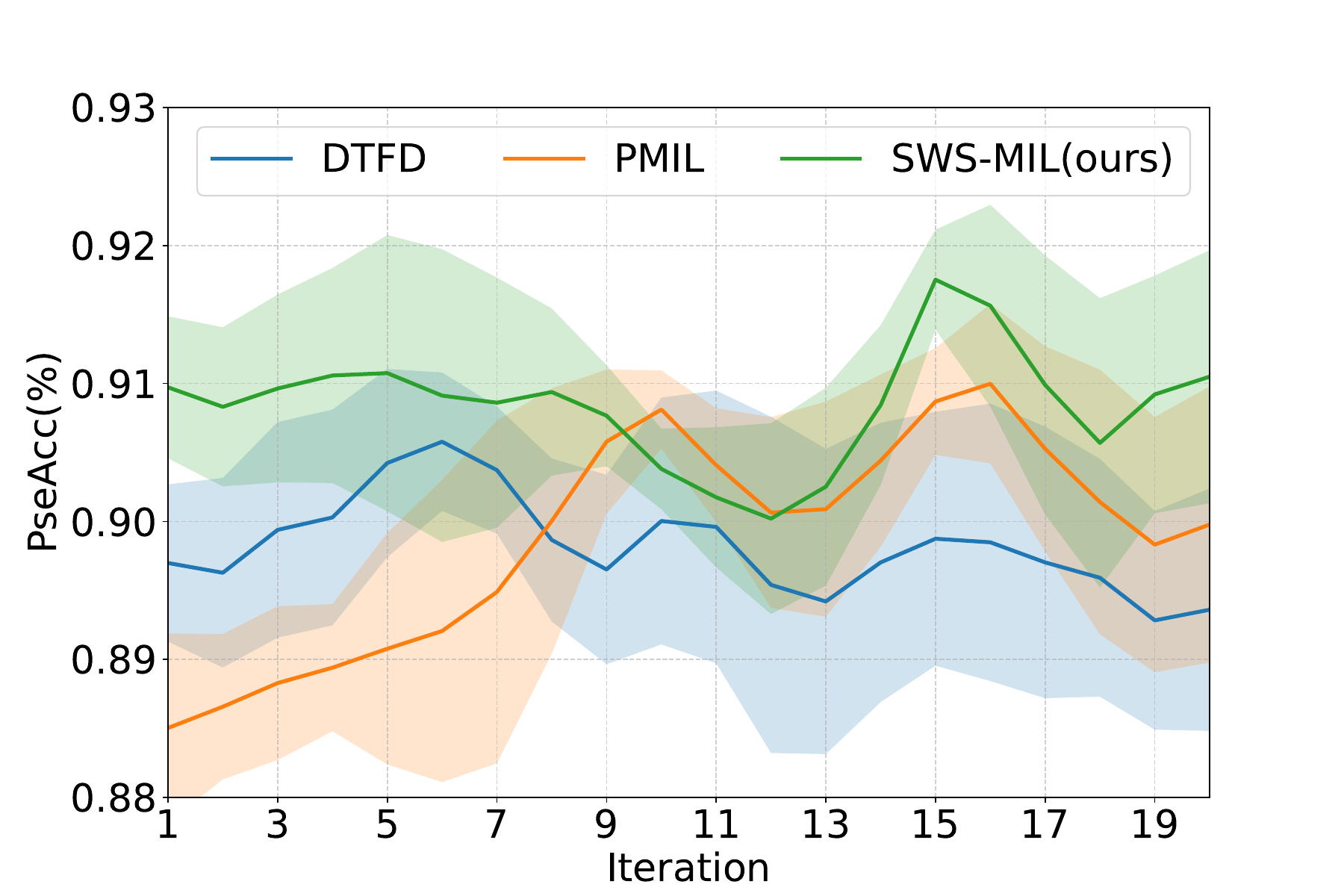}
\caption{Pseudo label accuracy $PseAcc$ of different pseudo bag augmentation methods during training.}
\label{acc}
\end{figure}
\textbf{TCGA-LUNG} includes 1,034 WSIs, comprising 528 cases of lung adenocarcinoma (LUAD) and 506 cases of lung squamous cell carcinoma (LUSC). We employ a 3-fold cross-validation protocol for both training and testing.

We utilized one-versus-others slide-level metrics: accuracy (ACC), the area under the curve (AUC), and macro F1 score (F1). We also introduced a new metric $PseAcc$ to measure the pseudo label accuracy of pseudo bag assignment.

\subsection{Implementation Details}
We tiled non-overlapping 256$\times$256 pixel patches, at the magnification of 5$\times$ for BRACS, and 20$\times$ for CAMELYON-16, and TCGA-LUNG, resulting in average counts of 7156, 714, and 11951 patches per bag, respectively.

All experiments were conducted on a workstation equipped with NVIDIA RTX 4090 GPUs. We employed ResNet50 pretrained from the ImageNet dataset as the encoder and ABMIL as the primary MIL model. We implemented an early stopping strategy with the patience parameter set to 10 epochs. The initial learning rate was set to 3e-4 and was then reduced to 1e-4. For the CAMELYON-16, BRACS, and TCGA-LUNG datasets, we set the maximum number of pseudo labels / bags to 8 / 4, 10 / 6, and 6 / 4.

The total number of training rounds was set to 10. We employed the same IIS estimation configuration for progressive pseudo bag augmentation \cite{yan2023shapley}.

\begin{figure*}[ht!]
\centering
\includegraphics[width=1\linewidth]{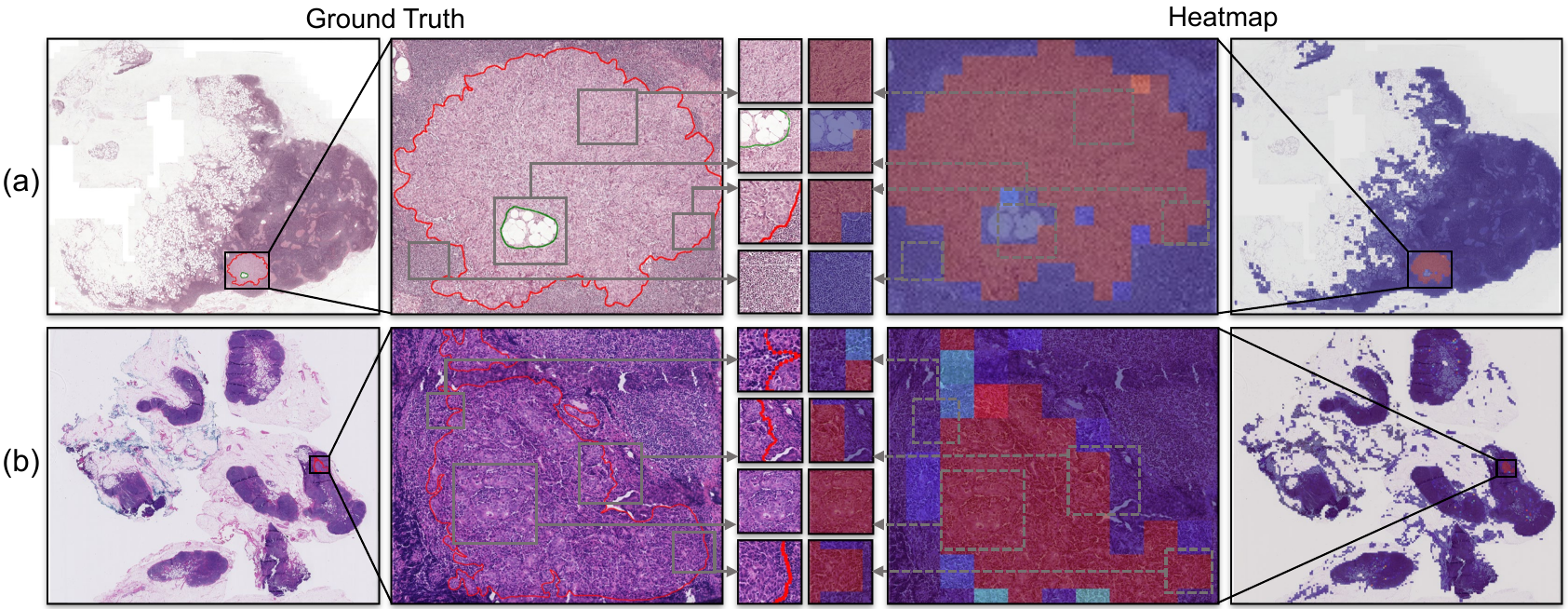}
\caption{Heatmap of our method in the CAMELYON-16 dataset. (a) and (b) are macro and micro metastasis cases. In the column of "Ground Truth", cancer and non-cancer regions are delineated in red and green, respectively. The "Heatmap" column represents the prediction results of SWS-MIL, where a redder color indicates greater importance of instance.}
\label{fig4}
\end{figure*}
\begin{figure*}[ht!]
\centering
\includegraphics[width=1\linewidth]{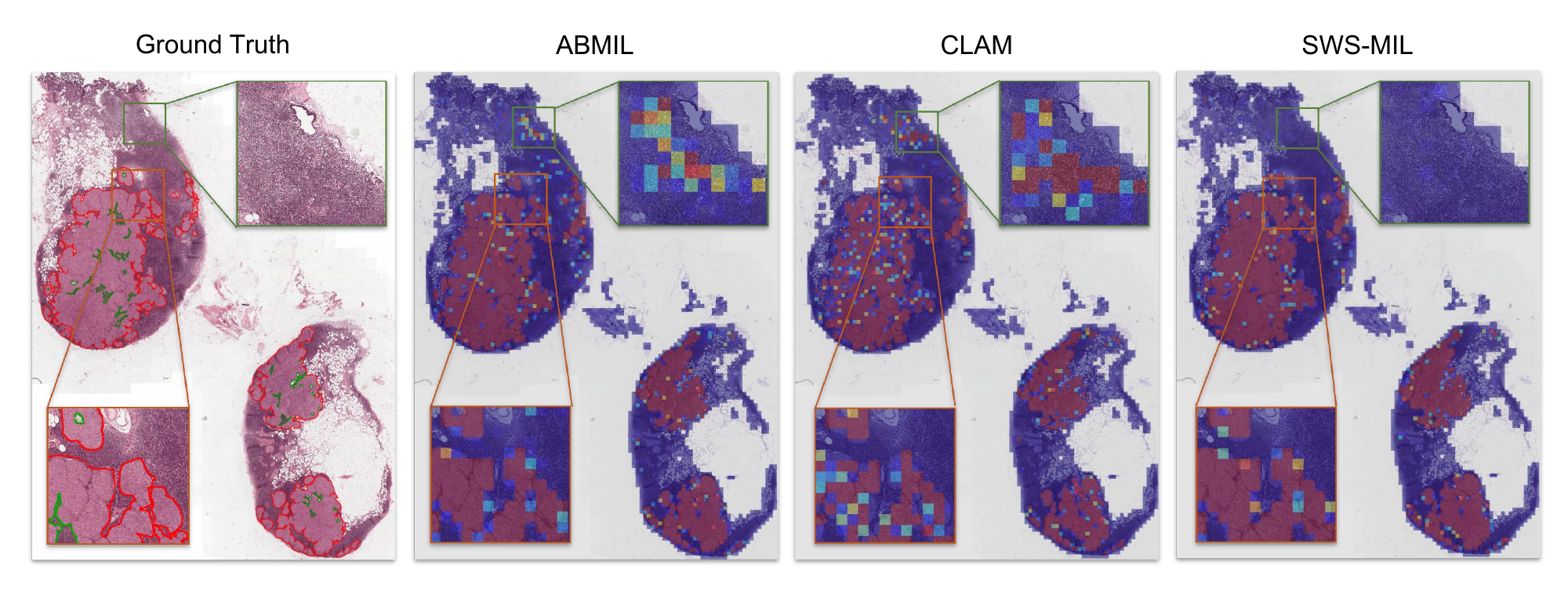}
\caption{Heatmaps of a case from CAMELYON-16 generated by ABMIL, CLAM, and SWS-MIL respectively. In the "Ground Truth" column, tumor / normal regions are delineated by red / green lines. Brighter red colors indicate higher tumor probabilities.}
\label{fig5}
\end{figure*}

\begin{table*}[ht]
    \caption{Ablation experiments of various augmentations on CAMELYON-16, BRACS and TCGA-LUNG. The subscripts are the standard deviation of each metric. The best evaluation results are in bold.}
    \centering
    \resizebox{\linewidth}{!}{
    \begin{tabular}{ccccccccccc} 
    \toprule
    \multicolumn{2}{c}{\textbf{Augmentation}} & \multicolumn{3}{c}{\textbf{CAMELYON-16}} & \multicolumn{3}{c}{\textbf{BRACS}} & \multicolumn{3}{c}{\textbf{TCGA-LUNG}} \\
    \cmidrule(r){1-2}\cmidrule(r){3-5} \cmidrule(lr){6-8} \cmidrule(lr){9-11}
    Student & Teacher & ACC(\%) & AUC(\%) & F1(\%) & ACC(\%) & AUC(\%) & F1(\%) & ACC(\%) & AUC(\%) & F1(\%) \\
    \midrule
    / & / & 87.44 \textsubscript{2.80} & 89.70 \textsubscript{0.70} & 86.26 \textsubscript{2.70} 
    & 66.79\textsubscript{1.72} & 84.37\textsubscript{1.57} & 55.24\textsubscript{2.62}
    & 91.20 \textsubscript{1.00} & 96.40 \textsubscript{0.70} & 91.20 \textsubscript{1.10}\\
    / & AdaPse 
    & 88.11 \textsubscript{1.61} & 89.05 \textsubscript{4.06} & 86.92 \textsubscript{1.79} 
    & 67.54\textsubscript{3.42} & \textbf{85.94\textsubscript{2.36}} & 57.76\textsubscript{4.26}
    & 91.70 \textsubscript{1.00} & 97.10 \textsubscript{0.50} & 91.70 \textsubscript{1.00}\\
    % Gaussian noise& AdaPse & 
    % 89.92 \textsubscript{2.19} & 92.50 \textsubscript{0.76} & 88.80 \textsubscript{2.49} &73.13 \textsubscript{4.26} & 87.55 \textsubscript{2.94} & 59.30 \textsubscript{5.23}\\
    % MixUp& AdaPse & 
    % 89.92 \textsubscript{1.10} & 92.84 \textsubscript{0.35} & 88.03 \textsubscript{1.42} & 72.55 \textsubscript{6.68} & 86.18 \textsubscript{3.36} & 63.02\textsubscript{10.38}\\
    MergeUp & AdaPse 
    & \textbf{90.18 \textsubscript{1.60}} & \textbf{93.25 \textsubscript{0.70}} & \textbf{89.27 \textsubscript{1.70}} 
    & \textbf{69.80\textsubscript{1.50}} & 84.90\textsubscript{1.20} & \textbf{58.20\textsubscript{2.90}}  
    &\textbf{92.30 \textsubscript{0.70}} & \textbf{97.20 \textsubscript{0.50}} & \textbf{92.20 \textsubscript{0.70}}\\
    \bottomrule
    \end{tabular}
    }
    \label{components}
\end{table*}
\subsection{Evaluation and Comparison}

We compared our performance on three datasets with various MIL methods, including Mean-Pooling MIL, Max-Pooling MIL, ABMIL~\cite{ilse2018attention}, DSMIL~\cite{li2021dual}, CLAM~\cite{lu2021data}, TransMIL~\cite{shao2021transmil}, DTFD~\cite{zhang2022dtfd}, R$^2$\text{T}~\cite{tang2024feature}, \text{and PMIL}~\cite{yan2023shapley}. 

We also evaluated different feature-augmented methods, including MixUp~\cite{zhang2017mixup}, AddictiveMIL~\cite{javed2022additive}, RankMix~\cite{chen2023rankmix}, and PseMix~\cite{liu2024pseudo}.

As shown in Table~\ref{compare_with_other_MIL}, significant results were obtained in comparisons with both various MIL methods and different feature-augmented methods. In slide-level classification, SWS-MIL achieved AUC scores of 93.25\%, 84.90\%, and 97.00\%, along with ACC scores of 90.18\%, 69.80\%, and 91.60\% for the CAMELYON-16, BRACS, and TCGA-LUNG datasets, respectively, which outperformed other mainstream methods, highlighting the dynamic adaptability of our model.

To explain the inner mechanism for the high performance of our method, we reported the pseudo label accuracy of different pseudo bag assignment methods during training, as illustrated in Figure~\ref{acc}. It revealed that random splitting like DTFD introduces more noise than IIS-based splitting, where SWS-MIL performed better than PMIL as it alleviates the mislabeling issue to some extent.

\subsection{Visualization and Interpretation}
To evaluate the effectiveness of our framework, we analyzed the heatmap of SWS-MIL for both macro and micro metastasis from CAMELYON-16, as shown in Figure~\ref{fig4} (a) and (b). In macro metastasis, our approach effectively avoids focusing on irrelevant fat or non-cancerous regions, which aligns with clinical annotations. In micro metastasis, our model accurately excludes regions labeled as normal in the ground truth, indicating that the model learns to characterize data more precisely.

Unlike weakly supervised learning frameworks, our semi-weakly supervised learning framework better utilizes inter-category information and optimizes pseudo label accuracy. As shown in Figure~\ref{fig5}, SWS-MIL achieves more accurate and precise attention heatmaps compared with ABMIL and CLAM, especially in outline. Additionally, ABMIL and CLAM inevitably focus on some non-cancer regions while our model excludes nearly all non-cancer regions.

\subsection{Ablation Study}
\subsubsection{Feature Augmentation Strategies.}
We evaluated the effectiveness of MergeUp and AdaPse augmentation in subsequent training on CAMELYON-16, BRACS, and TCGA-LUNG datasets. For the "teacher" model, we adopt AdaPse as the "weak augmentation". For the "student" model, we adopt MergeUp as the "strong augmentation". 

From the ablation studies, we summarize two insights:

1. Adaptive confidence threshold is effective in improving pseudo bag labeling accuracy. As shown in Table \ref{components}, the performance of the model with AdaPse augmentation improved by 1\% on both datasets compared to the baseline, especially on challenging BRACS. By setting adaptively adjusted pseudo bag label delineation thresholds, the model can better exclude incorrect labels and retain correct ones. 

2. Hybrid augmentation strategy works out. We analyzed the impact of adding a feature augmentation method on model performance by comparing our proposed MergeUp with a hybrid effective feature augmentation strategy, as shown in Table \ref{components}. The inclusion of the MergeUp strategy led to performance improvements across nearly all metrics in the three datasets. Specifically, metrics for datasets CAMELYON-16 and BRACS improved by 2$\%$ compared to using only AdaPse, with dataset TCGA-LUNG also showing enhancements. The only exception was a slight decrease in the AUC metric for the dataset BRACS, and this may be attributed to the challenging nature of the BRACS dataset, which is sensitive to the setting of the pseudo bag.

\begin{figure}[t]
\centering
\includegraphics[width=0.95\linewidth]{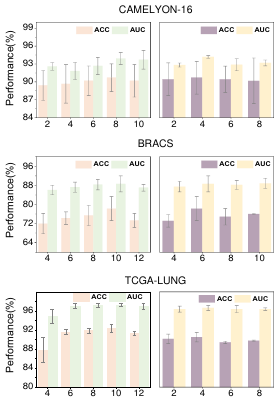}
\caption{Performance of SWS-MIL over different numbers of pseudo labels and bags on CAMELYON-16, BRACS, and TCGA-LUNG datasets.}
\label{baglabelnum}
\end{figure}

\subsubsection{Hyper-parameter Analysis.} 
As illustrated in Figure \ref{baglabelnum}, we evaluated two crucial hyper-parameters in SWS-MIL on the CAMELYON-16, BRACS, and TCGA-LUNG datasets: the number of pseudo bags / pseudo labels. Based on our experimental results, the most suitable hyper-parameters are 8 / 4 for CAMELYON-16, 10 / 6 for BRACS, and 10 / 4 for TCGA-LUNG.

% For the pseudo bag number, the result shows that as the number of bags increased, the performance of SWS-MIL initially improved on both datasets, reaching optimal values at 8 for CAMELYON-16 and 10 for BRACS, before subsequently declining. Similarly, the number of pseudo labels was assessed, revealing that model performance increased as the number of labels rose from 2 to 10, peaking at 4 for CAMELYON-16 and 6 for BRACS. 

% Our conclusions synthesize trends in ACC and AUC metrics. In the ablation experiments on the BRACS dataset, although the accuracy of SWS-MIL fluctuates smoothly as the number of pseudo bags increases, the AUC significantly improves when the number of pseudo bags is set to 10. Conversely, in the pseudo label number ablation experiment on the CAMELYON-16 dataset, the AUC metrics show minimal fluctuation as the number of pseudo labels increases. However, when the number of pseudo labels is set to 4, our model exhibits a significant advantage in ACC on CAMELYON-16.
For the number of pseudo bags, the results show that the performance of SWS-MIL initially improves on both datasets with the increase in the number of bags, reaching the best value of 8 on CAMELYON-16 and 10 on BRACS and TCGA-LUNG, respectively, and then decreases. Similarly, the number of pseudo labels is evaluated, and the results show that the model performance improves as the number of labels increases from 2 to 10, reaching a peak of 4 on CAMELYON-16 and TCGA-LUNG, and 6 on BRACS.

Our conclusion combines the trends of the ACC and AUC metrics. In the ablation experiments on the BRACS and TCGA-LUNG datasets, while the accuracy of SWS-MIL fluctuates smoothly with the increase in the number of pseudo bags, the ACC significantly improves when the number of pseudo bags is set to 10. In contrast, in the ablation experiments on the number of pseudo labels on the CAMELYON-16 dataset, the ACC metric shows minimal fluctuations with the increase in the number of pseudo labels. However, when the number of pseudo-labels is set to 4, our model shows a clear advantage in AUC on CAMELYON-16.
\section{Conclusion} 
Our method can present a generalized approach to further improve the efficiency of MIL models by enhancing the efficiency of inter-category feature learning and digging out more information without adding additional data and annotations to the datasets. It achieved top performance in the verification experiments across the three datasets CAMELYON-16, BRACS, and TCGA-LUNG. Our method also demonstrated strong characteristics in the ablation experiments, and based on these results, we selected the optimal hyperparameters. It is worth noting that our experimental report indicated that the inherent mechanism of our excellent performance is attributed to accurate pseudo label allocation.

In future work, we will continue to compress the module size to increase training speed. We will explore inter-category feature fusion methods for mutually exclusive datasets in the future.

\bibliography{ref}

\begin{thebibliography}{35}
\providecommand{\natexlab}[1]{#1}

\bibitem[{Avolio and Fuduli(2020)}]{avolio2020semiproximal}
Avolio, M.; and Fuduli, A. 2020.
\newblock A semiproximal support vector machine approach for binary multiple
  instance learning.
\newblock \emph{IEEE transactions on neural networks and learning systems},
  32(8): 3566--3577.

\bibitem[{Bernardini et~al.(2020)Bernardini, Morettini, Romeo, Frontoni, and
  Burattini}]{bernardini2020early}
Bernardini, M.; Morettini, M.; Romeo, L.; Frontoni, E.; and Burattini, L. 2020.
\newblock Early temporal prediction of type 2 diabetes risk condition from a
  general practitioner electronic health record: a multiple instance boosting
  approach.
\newblock \emph{Artificial Intelligence in Medicine}, 105: 101847.

\bibitem[{Borowsky et~al.(2020)Borowsky, Glassy, Wallace, Kallichanda, Behling,
  Miller, Oswal, Feddersen, Bakhtar, Mendoza et~al.}]{borowsky2020digital}
Borowsky, A.~D.; Glassy, E.~F.; Wallace, W.~D.; Kallichanda, N.~S.; Behling,
  C.~A.; Miller, D.~V.; Oswal, H.~N.; Feddersen, R.~M.; Bakhtar, O.~R.;
  Mendoza, A.~E.; et~al. 2020.
\newblock Digital whole slide imaging compared with light microscopy for
  primary diagnosis in surgical pathology: a multicenter, double-blinded,
  randomized study of 2045 cases.
\newblock \emph{Archives of pathology \& laboratory medicine}, 144(10):
  1245--1253.

\bibitem[{Brancati et~al.(2022)Brancati, Anniciello, Pati, Riccio,
  Scognamiglio, Jaume, De~Pietro, Di~Bonito, Foncubierta, Botti
  et~al.}]{brancati2022bracs}
Brancati, N.; Anniciello, A.~M.; Pati, P.; Riccio, D.; Scognamiglio, G.; Jaume,
  G.; De~Pietro, G.; Di~Bonito, M.; Foncubierta, A.; Botti, G.; et~al. 2022.
\newblock Bracs: A dataset for breast carcinoma subtyping in h\&e histology
  images.
\newblock \emph{Database}, 2022: baac093.

\bibitem[{Carbonneau, Granger, and Gagnon(2016)}]{carbonneau2016witness}
Carbonneau, M.-A.; Granger, E.; and Gagnon, G. 2016.
\newblock Witness identification in multiple instance learning using random
  subspaces.
\newblock In \emph{2016 23rd International Conference on Pattern Recognition
  (ICPR)}, 3639--3644. IEEE.

\bibitem[{Chen, Bi, and Wang(2006)}]{chen2006miles}
Chen, Y.; Bi, J.; and Wang, J.~Z. 2006.
\newblock MILES: Multiple-instance learning via embedded instance selection.
\newblock \emph{IEEE transactions on pattern analysis and machine
  intelligence}, 28(12): 1931--1947.

\bibitem[{Chen and Lu(2023)}]{chen2023rankmix}
Chen, Y.-C.; and Lu, C.-S. 2023.
\newblock Rankmix: Data augmentation for weakly supervised learning of
  classifying whole slide images with diverse sizes and imbalanced categories.
\newblock In \emph{Proceedings of the IEEE/CVF Conference on Computer Vision
  and Pattern Recognition}, 23936--23945.

\bibitem[{Gang, Yuan, and Bing(2013)}]{gang2013medical}
Gang, J.; Yuan, F.; and Bing, Z. 2013.
\newblock Medical image semantic annotation based on MIL.
\newblock In \emph{2013 ICME International Conference on Complex Medical
  Engineering}, 85--90. IEEE.

\bibitem[{Ilse, Tomczak, and Welling(2018)}]{ilse2018attention}
Ilse, M.; Tomczak, J.; and Welling, M. 2018.
\newblock Attention-based deep multiple instance learning.
\newblock In \emph{International conference on machine learning}, 2127--2136.
  PMLR.

\bibitem[{Javed et~al.(2022)Javed, Juyal, Padigela, Taylor-Weiner, Yu, and
  Prakash}]{javed2022additive}
Javed, S.~A.; Juyal, D.; Padigela, H.; Taylor-Weiner, A.; Yu, L.; and Prakash,
  A. 2022.
\newblock Additive mil: Intrinsically interpretable multiple instance learning
  for pathology.
\newblock \emph{Advances in Neural Information Processing Systems}, 35:
  20689--20702.

\bibitem[{Jia et~al.(2017)Jia, Huang, Eric, Chang, and Xu}]{jia2017constrained}
Jia, Z.; Huang, X.; Eric, I.; Chang, C.; and Xu, Y. 2017.
\newblock Constrained deep weak supervision for histopathology image
  segmentation.
\newblock \emph{IEEE transactions on medical imaging}, 36(11): 2376--2388.

\bibitem[{Komura and Ishikawa(2018)}]{komura2018machine}
Komura, D.; and Ishikawa, S. 2018.
\newblock Machine learning methods for histopathological image analysis.
\newblock \emph{Computational and structural biotechnology journal}, 16:
  34--42.

\bibitem[{Kumar, Gupta, and Gupta(2020)}]{kumar2020whole}
Kumar, N.; Gupta, R.; and Gupta, S. 2020.
\newblock Whole slide imaging (WSI) in pathology: current perspectives and
  future directions.
\newblock \emph{Journal of digital imaging}, 33(4): 1034--1040.

\bibitem[{Laine and Aila(2016)}]{laine2016temporal}
Laine, S.; and Aila, T. 2016.
\newblock Temporal ensembling for semi-supervised learning.
\newblock \emph{arXiv preprint arXiv:1610.02242}.

\bibitem[{Li, Li, and Eliceiri(2021)}]{li2021dual}
Li, B.; Li, Y.; and Eliceiri, K.~W. 2021.
\newblock Dual-stream multiple instance learning network for whole slide image
  classification with self-supervised contrastive learning.
\newblock In \emph{Proceedings of the IEEE/CVF conference on computer vision
  and pattern recognition}, 14318--14328.

\bibitem[{Liu et~al.(2024)Liu, Ji, Zhang, and Ye}]{liu2024pseudo}
Liu, P.; Ji, L.; Zhang, X.; and Ye, F. 2024.
\newblock Pseudo-Bag Mixup Augmentation for Multiple Instance Learning-Based
  Whole Slide Image Classification.
\newblock \emph{IEEE Transactions on Medical Imaging}.

\bibitem[{Lu et~al.(2021)Lu, Williamson, Chen, Chen, Barbieri, and
  Mahmood}]{lu2021data}
Lu, M.~Y.; Williamson, D.~F.; Chen, T.~Y.; Chen, R.~J.; Barbieri, M.; and
  Mahmood, F. 2021.
\newblock Data-efficient and weakly supervised computational pathology on
  whole-slide images.
\newblock \emph{Nature biomedical engineering}, 5(6): 555--570.

\bibitem[{Raciti et~al.(2023)Raciti, Sue, Retamero, Ceballos, Godrich, Kunz,
  Casson, Thiagarajan, Ebrahimzadeh, Viret et~al.}]{raciti2023clinical}
Raciti, P.; Sue, J.; Retamero, J.~A.; Ceballos, R.; Godrich, R.; Kunz, J.~D.;
  Casson, A.; Thiagarajan, D.; Ebrahimzadeh, Z.; Viret, J.; et~al. 2023.
\newblock Clinical validation of artificial intelligence--augmented pathology
  diagnosis demonstrates significant gains in diagnostic accuracy in prostate
  cancer detection.
\newblock \emph{Archives of Pathology \& Laboratory Medicine}, 147(10):
  1178--1185.

\bibitem[{Shao et~al.(2021)Shao, Bian, Chen, Wang, Zhang, Ji
  et~al.}]{shao2021transmil}
Shao, Z.; Bian, H.; Chen, Y.; Wang, Y.; Zhang, J.; Ji, X.; et~al. 2021.
\newblock Transmil: Transformer based correlated multiple instance learning for
  whole slide image classification.
\newblock \emph{Advances in neural information processing systems}, 34:
  2136--2147.

\bibitem[{Silva-Rodr{\'\i}guez, Colomer, and Naranjo(2021)}]{silva2021weglenet}
Silva-Rodr{\'\i}guez, J.; Colomer, A.; and Naranjo, V. 2021.
\newblock WeGleNet: A weakly-supervised convolutional neural network for the
  semantic segmentation of Gleason grades in prostate histology images.
\newblock \emph{Computerized Medical Imaging and Graphics}, 88: 101846.

\bibitem[{Sohn et~al.(2020)Sohn, Berthelot, Carlini, Zhang, Zhang, Raffel,
  Cubuk, Kurakin, and Li}]{sohn2020fixmatch}
Sohn, K.; Berthelot, D.; Carlini, N.; Zhang, Z.; Zhang, H.; Raffel, C.~A.;
  Cubuk, E.~D.; Kurakin, A.; and Li, C.-L. 2020.
\newblock Fixmatch: Simplifying semi-supervised learning with consistency and
  confidence.
\newblock \emph{Advances in neural information processing systems}, 33:
  596--608.

\bibitem[{Tang et~al.(2024)Tang, Zhou, Huang, Zhu, Zhang, and
  Liu}]{tang2024feature}
Tang, W.; Zhou, F.; Huang, S.; Zhu, X.; Zhang, Y.; and Liu, B. 2024.
\newblock Feature Re-Embedding: Towards Foundation Model-Level Performance in
  Computational Pathology.
\newblock \emph{arXiv preprint arXiv:2402.17228}.

\bibitem[{Tarvainen and Valpola(2017)}]{tarvainen2017mean}
Tarvainen, A.; and Valpola, H. 2017.
\newblock Mean teachers are better role models: Weight-averaged consistency
  targets improve semi-supervised deep learning results.
\newblock \emph{Advances in neural information processing systems}, 30.

\bibitem[{Wang et~al.(2024)Wang, Shi, Yan, Sun, Zhu, Guan, and
  He}]{wang2024task}
Wang, X.; Shi, S.; Yan, R.; Sun, Q.; Zhu, L.; Guan, T.; and He, Y. 2024.
\newblock Task-oriented Embedding Counts: Heuristic Clustering-driven Feature
  Fine-tuning for Whole Slide Image Classification.
\newblock \emph{arXiv preprint arXiv:2406.00672}.

\bibitem[{Wang et~al.(2018)Wang, Yan, Tang, Bai, and Liu}]{wang2018revisiting}
Wang, X.; Yan, Y.; Tang, P.; Bai, X.; and Liu, W. 2018.
\newblock Revisiting multiple instance neural networks.
\newblock \emph{Pattern Recognition}, 74: 15--24.

\bibitem[{Wang et~al.(2022)Wang, Chen, Heng, Hou, Fan, Wu, Wang, Savvides,
  Shinozaki, Raj et~al.}]{wang2022freematch}
Wang, Y.; Chen, H.; Heng, Q.; Hou, W.; Fan, Y.; Wu, Z.; Wang, J.; Savvides, M.;
  Shinozaki, T.; Raj, B.; et~al. 2022.
\newblock Freematch: Self-adaptive thresholding for semi-supervised learning.
\newblock \emph{arXiv preprint arXiv:2205.07246}.

\bibitem[{Xie et~al.(2020)Xie, Dai, Hovy, Luong, and Le}]{xie2020unsupervised}
Xie, Q.; Dai, Z.; Hovy, E.; Luong, T.; and Le, Q. 2020.
\newblock Unsupervised data augmentation for consistency training.
\newblock \emph{Advances in neural information processing systems}, 33:
  6256--6268.

\bibitem[{Yan et~al.(2023)Yan, Sun, Jin, Liu, He, Guan, and
  Chen}]{yan2023shapley}
Yan, R.; Sun, Q.; Jin, C.; Liu, Y.; He, Y.; Guan, T.; and Chen, H. 2023.
\newblock Shapley Values-enabled Progressive Pseudo Bag Augmentation for Whole
  Slide Image Classification.
\newblock \emph{arXiv preprint arXiv:2312.05490}.

\bibitem[{Yang et~al.(2023)Yang, He, Jiao, Pan, Wang, Wang, and
  Wu}]{yang2023multiple}
Yang, B.; He, Y.; Jiao, C.; Pan, X.; Wang, G.; Wang, L.; and Wu, J. 2023.
\newblock Multiple Instance Metric Learning Network for Hyperspectral Target
  Detection.
\newblock \emph{IEEE Transactions on Geoscience and Remote Sensing}.

\bibitem[{Yang et~al.(2022)Yang, Chen, Zhao, Yang, Zhang, He, and
  Yao}]{yang2022remix}
Yang, J.; Chen, H.; Zhao, Y.; Yang, F.; Zhang, Y.; He, L.; and Yao, J. 2022.
\newblock Remix: A general and efficient framework for multiple instance
  learning based whole slide image classification.
\newblock In \emph{International Conference on Medical Image Computing and
  Computer-Assisted Intervention}, 35--45. Springer.

\bibitem[{Yu et~al.(2023)Yu, Wu, Ming, Deng, Li, Ou, He, Wang, Zhang, and
  Wang}]{yu2023prototypical}
Yu, J.-G.; Wu, Z.; Ming, Y.; Deng, S.; Li, Y.; Ou, C.; He, C.; Wang, B.; Zhang,
  P.; and Wang, Y. 2023.
\newblock Prototypical multiple instance learning for predicting lymph node
  metastasis of breast cancer from whole-slide pathological images.
\newblock \emph{Medical Image Analysis}, 85: 102748.

\bibitem[{Zhang et~al.(2017)Zhang, Cisse, Dauphin, and
  Lopez-Paz}]{zhang2017mixup}
Zhang, H.; Cisse, M.; Dauphin, Y.~N.; and Lopez-Paz, D. 2017.
\newblock mixup: Beyond empirical risk minimization.
\newblock \emph{arXiv preprint arXiv:1710.09412}.

\bibitem[{Zhang et~al.(2022)Zhang, Meng, Zhao, Qiao, Yang, Coupland, and
  Zheng}]{zhang2022dtfd}
Zhang, H.; Meng, Y.; Zhao, Y.; Qiao, Y.; Yang, X.; Coupland, S.~E.; and Zheng,
  Y. 2022.
\newblock Dtfd-mil: Double-tier feature distillation multiple instance learning
  for histopathology whole slide image classification.
\newblock In \emph{Proceedings of the IEEE/CVF Conference on Computer Vision
  and Pattern Recognition}, 18802--18812.

\bibitem[{Zheng et~al.(2024)Zheng, Wang, Zhu, Yan, Li, Wei, Zhang, Du, Guo, He
  et~al.}]{zheng2024deep}
Zheng, R.; Wang, X.; Zhu, L.; Yan, R.; Li, J.; Wei, Y.; Zhang, F.; Du, H.; Guo,
  L.; He, Y.; et~al. 2024.
\newblock A Deep Learning Method for Predicting The Origins of Cervical Lymph
  Node Metastatic Cancer on Digital Pathological Images.
\newblock \emph{iScience}.

\bibitem[{Zhu et~al.(2023)Zhu, Shi, Wei, Wang, Shi, Zhang, Yan, Liu, He, Wang
  et~al.}]{zhu2023accurate}
Zhu, L.; Shi, H.; Wei, H.; Wang, C.; Shi, S.; Zhang, F.; Yan, R.; Liu, Y.; He,
  T.; Wang, L.; et~al. 2023.
\newblock An accurate prediction of the origin for bone metastatic cancer using
  deep learning on digital pathological images.
\newblock \emph{EBioMedicine}, 87.

\end{thebibliography}

\end{document}